\documentclass[aps,11pt,final,notitlepage,oneside,onecolumn,nobibnotes,nofootinbib,superscriptaddress,
noshowpacs,centertags]{revtex4}
\usepackage{graphicx}
\usepackage{amsfonts}
\newcommand{\R}{\mathbb{R}}

\usepackage{algorithm}
\usepackage{algorithmic}

\DeclareMathAlphabet{\mathpzc}{OT1}{pzc}{m}{it}
\newcommand{\db}{\mathpzc{d}_\mathcal{B}}
\newcommand{\bcal}{\mathcal{B}}

\begin{document}

\title{Fast, Linear Time, m-Adic Hierarchical Clustering for Search and Retrieval 
using the Baire Metric, with linkages to Generalized Ultrametrics, Hashing, 
Formal Concept Analysis, and Precision of Data Measurement}

\author{Fionn Murtagh (1,3) and Pedro Contreras (2,3) \\
(1) Science Foundation Ireland, Wilton Park House, Wilton Place, Dublin 2, Ireland \\
(2) Thinking Safe Ltd., Royal Holloway University of London, Egham TW20 0EX, England \\
(3) Department of Computer Science, Royal Holloway University of London, Egham TW20 0EX, England \\
fmurtagh@acm.org, pedro.contreras@thinkingsafe.com}

\begin{abstract}
We describe many vantage points on the Baire metric and its use in clustering 
data, or its use in preprocessing and structuring data in order to support 
search and retrieval operations.   In some cases, we proceed directly to 
clusters and do not directly determine the distances.   We show how 
a hierarchical clustering can be read directly from one pass through the data.
We offer insights also on practical implications of precision of data 
measurement.   As a mechanism for treating multidimensional data, including 
very high dimensional data, we use random projections.   
\end{abstract}

\maketitle

\section{Introduction}

In areas such as search, matching, retrieval and general data analysis, 
massive increase in data requires new methods that can cope well 
with the explosion in volume and dimensionality of the available data. 
In this work, the Baire metric, which is furthermore an ultrametric, is used 
to induce a hierarchy and in turn to support clustering, matching and other operations. 

Arising directly out of the Baire distance is an ultrametric tree, which also can be seen as a tree that hierarchically clusters data. This  presents a number of advantages when storing and retrieving data. When the data source is in numerical form this ultrametric tree can be used as an index structure making matching and search, and thus retrieval, much easier. 

The clusters can be associated with hash keys, that is to say, the cluster members 
can be mapped onto ``bins'' or ``buckets''.    

Another vantage point in this work is precision of measurement.  Data measurement precision can be
either used as given or modified in order to enhance the inherent ultrametric 
and hence hierarchical properties of the data.  

Rather than mapping pairwise relationships onto the reals, as distance does, we
can alternatively map onto subsets of the power set of, say, attributes of our 
observation set.  This is expressed by the generalized ultrametric, which 
maps pairwise relationships into a partially ordered set.  It is also current practice 
as formal concept analysis where the range of the mapping is a lattice.   

Relative to other algorithms the Baire-based hierarchical clustering method is fast.
It is a direct reading algorithm involving one scan of the input data set, and
is of linear computational complexity. 

Many vantage points are possible, all in the Baire metric framework.  The following 
vantage points will be discussed in this article.   

\begin{itemize}
\item Metric that is simultaneously an ultrametric.
\item Hierarchy induced through m-adic encoding (m positive integer, e.g.\ 10).
\item p-Adic (p prime) or m-adic clustering.

\item Hashing of data into bins.
\item Data precision of measurement implies
how hierarchical the data is.

\item Generalized ultrametric.
\item Lattice-based formal concept analysis.
\item Linear computational time hierarchical clustering.  
\end{itemize}

\section{The Baire Metric, the Baire Ultrametric}

\subsection{Metric and Ultrametric Spaces}\label{subsection:metric-ultrametric}

Our purpose  consists of mapping data into an ultrametric space or, alternatively
expressed, searching 
for an ultrametric embedding, or ultrametrization~\cite{Rooij78}. Actually,  
inherent ultrametricity leads to an identical result relative to 
most commonly used 
agglomerative criteria~\cite{Murtagh85-1}. Furthermore, data coding can help 
greatly in finding how inherently ultrametric data is~\cite{Murtagh04},
and this is further discussed in section \ref{sect6}.

A metric space $(X,d)$ consists of a set~$X$ on which is defined a 
\emph{distance function} $d$ which assigns to each pair of points of 
$X$ a distance 
between them, and satisfies the following four axioms for any triplet of 
points 
$x, y, z$:

\begin{enumerate}
        \item $\forall x, y \in X,  d(x,y) \geq 0$ (positiveness);
        \item $\forall x, y \in X, d(x,y) = 0$ \emph{iff} $x = y$ 
(reflexivity);
        \item $\forall x, y \in X, d(x,y) = d(y,x)$  (symmetry);
        \item $\forall x, y, z \in X, d(x,z) \leq d(x,y) + d(y,z)$  
(triangle inequality).
\end{enumerate}

An \emph{ultrametric space} respects the 
``strong triangular inequality'' or \emph{ultrametric inequality} defined as:
\[d(x,z) \leq max~\{d(x,y), \ d(y,z)\},\]
in addition to the positivity, reflexivity and symmetry properties 
for any triplet of points  $x, y, z \in X$.

Various properties of an ultrametric space ensue from this.  
For example, the triangle formed by any triplet is necessarily 
isosceles, with the two large sides equal; or is equilateral. Every point 
of a circle 
in an ultrametric space is a center of the circle. Two circles of the same 
radius, that are not disjoint, are overlapping~\cite{khrennikov,Lerman81}. 
Additionally, 
an ultrametric is a distance that is defined strictly on a tree, which is a 
property that is very useful in classification \cite{benz}.

\subsection{Ultrametric Baire Space and Distance}\label{subsection:baire}

A Baire space consists of countably infinite sequences with a metric defined 
in terms of the longest common prefix: the longer the common prefix, the 
closer a pair of sequences. What is of interest to us is this longest 
common 
prefix metric, which we call the Baire distance~\cite{bradley,Mikin79,Murtagh07}.

We begin with the 
longest common prefixes at issue being digits of precision of 
univariate or scalar values. 
 For example, let us consider two such decimal
values, $x$ and $y$, with both measured to some maximum precision.  One or 
both can be padded with 0s to have this maximum precision.
With no loss of generality we take $x$ and $y$ to be bounded by 0 and 1. 
Thus we consider ordered sets $x_{k}$ and $y_{k}$ for $k \in K$. 
So $k = 1$ is the first decimal place of precision; 
$k = 2$ is the second decimal place; . . . ; $k = \left| K \right|$ is 
the $\left| K 
\right|th$ decimal place.  The cardinality of the set K is the precision with 
which a number, $x$ or $y$, is measured.

Consider as examples $x_{3} = 0.478$; and $y_{3} = 0.472$. Start from the 
first decimal position.
For $k = 1$, we find $x_{1} = y_{1} = 4$. For $k = 2$, $x_{2} = y_{2} = 7$. 
But for $k = 3$,  $x_{3} \neq y_{3}$.

We now introduce the following distance (case of vectors $x$ and $y$, with 
1 attribute, hence unidimensional):

\begin{equation}
\label{eq:baire}
\db(x_{K}, y_{K}) =
        \left\{
        \begin{array}{ll}
       1 &\;\; $if$\;\;  x_{1} \neq y_{1}\\
       $inf$\;\;  \bcal^{-n} & \;\;\;\;\;\; x_{n} = y_{n}, \;\;\; 1 \leq n \leq
       \left| K \right|
    \end{array}
    \right.
\end{equation}

We call this $\db$ value Baire distance, which is a 1-bounded 
ultrametric~\cite{bradley,Murtagh07} distance, $0 < \db \leq 1$.
When dealing with binary (boolean) 
data 2 is the chosen base, $\bcal = 2$. When working with real
numbers the base is best defined to be 10, $\bcal = 10$.
With $\bcal = 10$, for instance, it can be seen that the Baire 
distance is embedded in a 10-way tree which leads to a convenient 
data structure to support search and other operations when we have 
decimal data.  As a consequence data can be organized, stored and 
accessed very efficiently and effectively in such a tree.  

For $\bcal$ prime, this distance has been studied by Benois-Pineau
et al.\ \cite{benois} 
and by Bradley \cite{bradley2009}, with many further (topological 
and number theoretic, leading to algorithmic and computational) 
insights arising from the 
p-adic (where p is prime) framework.

\section{Multidimensional Use of the Baire Metric through Random Projections}
\label{sectrp}

It is well known that traditional clustering methods do not scale 
well in very high dimensional spaces. A standard and widely used approach 
when dealing with high dimensionality is to apply a dimensionality reduction 
technique. This consists of finding a mapping $F$ relating the input data 
from the space $\mathbb{R}^{d}$ to a lower-dimension feature space 
$\mathbb{R}^{k}$:  
$F :\mathbb{R}^{d} \rightarrow \mathbb{R}^{k}$.

A least squares optimal way of reducing dimensionality is to project the 
data onto a lower dimensional orthogonal subspace. Principal component 
analysis (PCA) is a popular choice to do this.  It uses a linear 
transformation to form a simplified (viz.\ reduced dimensionality), 
dataset while retaining the characteristics (viz.\ variances) 
of the original data. PCA selects a best fitting, ordered sequence of subspaces
(of dimensionality 1, 2, 3, $\dots$) that best preserve the variance of the data.

This is a good solution when the data allows these calculations, but PCA as 
well as other dimensionality reduction techniques remain expensive from a
computational point of view, for very large data sets.  
 The essential eigenvalue and eigenvector 
decomposition is of $O(d^3)$ computational complexity.  Looking beyond the
least squares and orthogonal PCA projection, we have studied the benefits of
random projections.   

Random projection~\cite{Bingham01, Dasgupta06, Deegalla06, Fern07, Fradkin03, 
Li06, Lin03, Vempala04} is the finding of a low dimensional embedding of a 
point set, such that the distortion of any pair of points is bounded by a 
function of the lower dimensionality.

The theoretical support for random projection can be found in the 
Johnson-Lindenstrauss Lemma~\cite{Johnson84}. It states that a set of 
points in a high 
dimensional Euclidean space can be projected into a low dimensional Euclidean 
space such that the distance between any two points changes by a fraction of 
$1 + \varepsilon$, where $\varepsilon \in (0,1)$.

\medskip

\noindent
{\bf Johnson-Lindenstrauss Lemma}

\smallskip

\noindent
For any $0 < \varepsilon < 1$ and any integer $n$, let $k$ be a positive 
integer such that
        \begin{equation}
                k \geq 4(\varepsilon^{2}/2 - \varepsilon^{3}/3)^{-1} \ln n  
        \end{equation}
Then for any set $V$ of any points in $\mathbb{R}^{d}$, there is a map 
$f:\mathbb{R}^{d} \rightarrow \mathbb{R}^{k}$ such that for all $u$, $v \in V$,

$$(1 - \varepsilon) \parallel u - v \parallel^{2} ~ \leq ~ \parallel f(u) - 
f(v) \parallel^{2} ~ \leq ~ (1 + \varepsilon) \parallel u - v \parallel^{2}$$.

Furthermore, this map can be found in randomized polynomial time.
\medskip

The original proof \cite{Johnson84} 
was further simplified by Frankl and Maehara~\cite{Frankl88}, and 
Dasgupta and Gupta~\cite{Dasgupta03}.  See also \cite{Achlioptas01,Vempala04}.

In practice we find that random directions of high dimensional vectors 
are a sufficiently good approximation to an orthogonal system of axes. 
We present experimental results in \cite{pedrophd}.  
In this way we can exploit data sparsity in high dimensional spaces.

In random projection the original $d$-dimensional data is projected 
to a $k$-dimensional subspace $(k << d)$, using a random $k \times d$ 
matrix $R$:

\begin{equation}\label{eq:random-projection}
        X^\prime_{k \times N} = R_{k \times d}~X_{d \times N}
\end{equation}
where $X_{d \times N}$ is the original set with $d$-dimensionality and $N$ 
observations.
From the computational aspect, forming the random 
matrix $R$ and projecting the $d \times N$ data matrix $X$ into the $k$ 
dimensions is of order $O(dkN)$. If $X$ is sparse with $c$ non-zero 
entries per column, the complexity is of order $O(ckN)$.

\begin{figure}
  \begin{center}
  \includegraphics[scale=0.85]{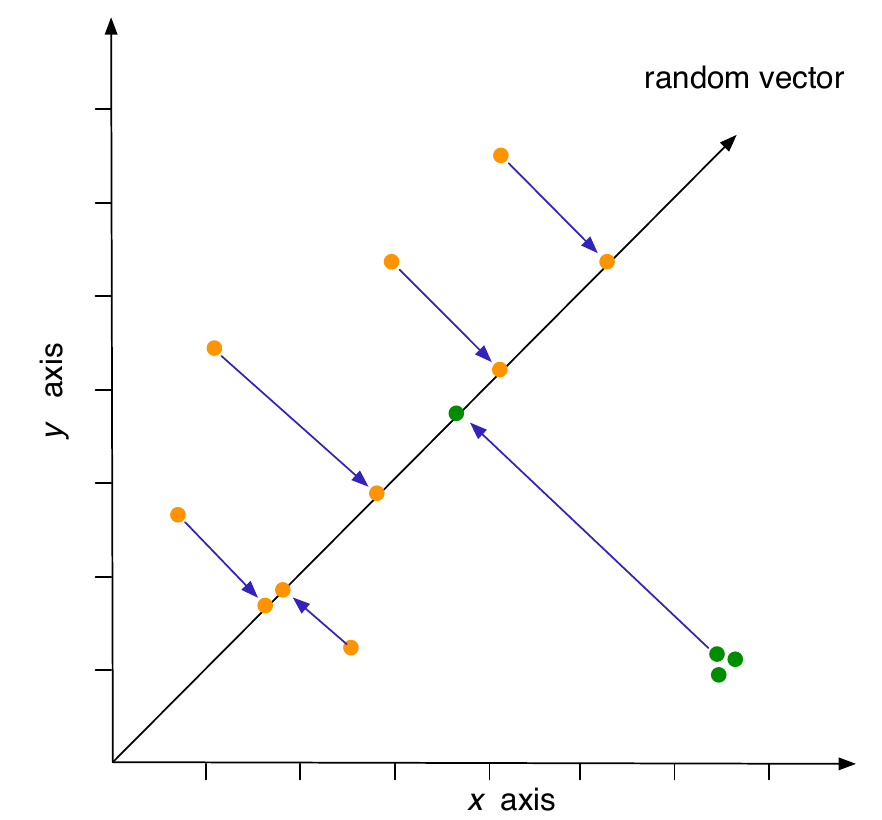}
    \caption{Random projection axis, showing orthogonal projections.}  
  \label{fig:rp}
  \end{center}
\end{figure}

Random projection can be seen as a class of hashing function.
(This is further discussed in section \ref{hashing} below.)  
Hashing is much faster than alternative methods because it avoids the 
pairwise 
comparisons required for partitioning and classification. This process is 
depicted 
in a Euclidean 2-dimensional space in Figure~\ref{fig:rp}, where a random 
vector is drawn and data points are projected onto it. If two points $(p, q)$ are 
close, they will have a very small $\| p-q \|$ (Euclidean metric) value; and they 
will hash to the same value with high probability. If they are distant, they 
should collide with small probability.

\section{Hierarchical Tree Defined from m-Adic Encoding}
\label{sect4}

Consider data values, base 10, that are $> 0, < 1$.
Let the full data set be associated with the root of a regular 10-way tree.  
Determine 10 clusters/bins at the first level from the root of the tree, 
labeled through the first digit of precision, 0, 1, 2, ... , 9.
Determine the first level of the tree -- for each of the 10 first level 
clusters -- labeled through the second digit of precision.
The clusters/bins, associated with terminals in the tree, can be 
labeled 00, 01, 02, ... , 09; 10, 11, ... , 19; 20, 21, ... , 29; 
... ... ... 90, 91, 92, ... , 99.   
This m-adic encoding tree, with m = 10, can be continued through further levels.

In \cite{CMJoC}, a large cosmology survey data set is used, where
it is sought to match spectrometric redshifts against photometric 
redshifts (respectively denoted z\_spec and z\_phot).  
Redshift is (i) cosmological distance, (ii) recession
velocity, (iii) look-back time to the observation in question,
and (iv) the third dimension of the 3D cosmos in addition to 
(in the commonly used, extra-Solar System, coordinate frame) right ascension and declination.  
Using spectrometry is demanding in terms of instrumental procedure, 
but furnishes better quality output (in terms of precision and error).  
Photometric redshifts,
on the other hand, are more easily obtained but are of lower quality.
Hence there is interest in inferring spectrometric redshifts
from photometric redshifts.  With that aim in mind, in \cite{CMJoC} 
we looked at a clusterwise regression approach,
where ``clusterwise'' is not spatial but rather based on 
measurement precision.  

To summarize our results on approximately 400,000 pairs of 
redshift measurements, we found the following. 

\begin{itemize}
\item 82.8\% of z\_spec and z\_phot have at least 2 common prefix digits.

This relates to numbers of redshift couples 
sharing 6, 5, 4, 3, or 2 (precision-ordered) decimal digits. 

We can find very efficiently where these 82.8\% of the 
astronomical objects are, in our data.

\item 
21.7\% of z\_spec and z\_phot have at least 3 common prefix digits. 

This relates to numbers of observations sharing 6, 5, 4, or 3 decimal digits. 
\end{itemize}

This exemplifies how we read off clusters from the 
hierarchical tree furnished by the Baire (ultra)metric.  

\section{Longest Common Prefix and Hashing}
\label{hashing}

The longest common prefix, used in the Baire metric, can be 
viewed as a hashing or data binning scheme.  We will follow up 
first on the implications of this when used in tandem with
random projections for multivariate data (i.e., data in 
a Euclidean or other space of dimensionality $> 1$).  

\subsection{From Random Projection to Hashing}

Random projection
is the finding of a low dimensional embedding of a point set -- dimension
equals 1, hence a line or axis, in this work  -- such that the
distortion of any pair of points is bounded by a function of the lower
dimensionality \cite{Vempala04}.  As noted in subsection 
\ref{sectrp}, there is extensive literature in this
area, e.g.\ \cite{Dutta06}.
While random projection {\em per se} will not  guarantee
a bijection of best match in original and in lower dimensional spaces,
our use of projection here is effectively a hashing method.  We aim 
to deliberately find hash collisions that thereby provide
a sufficient condition for the mapped vectors to be matched.
Alternatively expressed, candidate best match vectors are determined 
in this way.    
As an example of this approach, Miller et al.\ \cite{Miller05}
use the the MD5 (Message Digest 5) hashing scheme as a basis for 
nearest neighbor searching.  Buaba et al.\ \cite{buaba} note that
hashing is an approximate matching approach whereby (i) the probability 
of not finding a nearest neighbor is very small, and (ii) neighbors
in the same hash class furnished by the hashing projection are good 
approximations to the (optimal) nearest neighbor.   Retrieval of 
audio data is at issue in \cite{Miller05}, retrieval of Earth 
observation data is studied in \cite{buaba}, and content-based
image retrieval is the focus of \cite{wu2}.  

A hash function, used for similarity finding, maps data into a 
fixed length data {\em key}.   Thus, possibly following random 
projection, assume our data includes two strings of values 
0.457891 and 0.457883456.  Now consider how both of these can be 
put into the same ``bin'' or ``bucket'' labeled by 4578.   In 
this case the fixed length hash key of a data value is read off 
as the first 4 significant digits.  

Collision of identically valued vectors is guaranteed, but what of
collision of non-identically valued vectors, which we want to avoid?
Such a result can be established based on the assumption of what 
distribution our original data follow.  A stable
distribution is used in \cite{indyk}, viz.\ a distribution such 
that a limited number of
weighted sums of the  variables is also itself of the same distribution.
Examples include both Gaussian (which is 2 stable, \cite{indyk2})
and power law (long tailed) distributions.

Interestingly, however, very high dimensional (or equivalently, very low
sample size or low $N$) data sets, by
virtue of high relative dimensionality alone, have points mostly lying
at the vertices of a regular simplex or polygon \cite{hall, Murtagh04}.  
Such regular sparsity is one reason why we have found random projection
to work well.  Another reason is that we use attribute weighting (e.g.\
\cite{Murtagh07}).   Li 
et al.\ \cite{li} (see their section 5) note that pairwise distance can 
be ``meaningless'' when using heavy tailed distributions but this problem
is bypassed by attribute weighting which modifies the data's 2nd and
higher order moments.  Thereafter, with the random projection mapping 
using statistically uniformly drawn 
weights for attribute $j$, $w_j$, then the random projections for data vectors 
$x$ and $y$ are respectively $\sum_j w_j x_j$ and
$\sum_j w_j x'_j$.  
We can anticipate near equal $x_j$ and $x'_j$ terms, for all $j$, to be mapped
onto fairly close resultant scalar values.

We adopted an experimental approach to confirm these 
hypotheses, viz., that high dimensional
data are ``regular'' or ``structured'' in such a way;
and that, as a consequence, hashing is particularly well-behaved
in the sense of non-identical vectors being nearly always
collision-free.  We studied stability of results, and also 
effectiveness relative to other clustering methods, in particular
k-means partitioning.   
In \cite{Murtagh07}, this 
principle of binning data is used on a large, high dimensional 
chemoinformatics data set.  
It is shown
in \cite{CMJoC} how a large astronomical data set also lends 
itself very well to similarity finding in this way.  

\section{Enhancing Ultrametricity through Precision of Measurement}
\label{sect6}

By reducing the precision of measurement we are in effect 
mapping the data into bins, or hashing the data.
Furthermore the longest common 
prefix as used in the Baire metric gives us one way, that is both
practical and useful, to extract reduced precision information
from our data. 

\subsection{Quantifying Ultrametricity}
 
We assume Euclidean space.  If necessary our data can be mapped into 
a Euclidean space, see e.g.\ \cite{Murtagh} that maps contingency table
data endowed with the chi squared metric into a Euclidean factor space.
In a different application domain where data are given as pairwise comparisons
or preferences, multidimensional scaling methods take pairwise ranks (hence
ordinal data) and perform a mapping into a Euclidean space.   

Consider, in Euclidean space, a triplet of points that defines a triangle.
Take the smallest internal angle, $a$, in triangle $\leq 60$ degrees.
For the two other internal angles, $b$ and $c$, if $| b - c | < 2$ 
degrees then we characterize the triangle as being approximately 
isosceles with small base, or equilateral.  That is to say, 
we consider 2 degrees to be an arbitrary small angle.   Because of the
approximation involved we could claim, informally, that this leads to a 
fuzzy definition of ultrametricity.  

Any triangle in Euclidean space is ultrametric if it is isosceles
with small base, or equilateral \cite{Lerman81}.
 
To use this in practice, we look for the overall proportion of 
such triangles in our data.  This yields a coefficient of ultrametricity
\cite{Murtagh04}.
Therefore to quantify ultrametricity we take all possible triplets, $i, j, k$.
We look at their angles, and judge whether or not the ultrametric 
triangle properties are verified.
Having examined all possible triangles, our ultrametricity measure that 
we term $\alpha$ is the 
number of ultrametric-verifying triangles, divided by the total 
number of triangles. 
When all triangles respect the ultrametric properties this yields 
$\alpha = 1$; if no triangle does, then $\alpha = 0$.
For $N$ objects, exhaustive enumeration of triangles is 
computationally prohibitive, so we 
sample $i,j,k$ in practice.  We sample uniformly over the object set.

At least two other approaches to quantifying ultrametrictiy 
have been used but in \cite{Murtagh04} we 
point to their limitations.   First, there is the 
relationship between subdominant ultrametric, and given dissimilarities.  
See \cite{Rammal86}.  
Secondly, we may look at 
whether interval between median and maximum rank dissimilarity of every set 
of triplets is nearly empty.  Taking ranks provides scale invariance. 
This is the approach of \cite{Lerman81}. 

\subsection{Ultrametricity is Pervasive}

We find experimentally 
that ultrametricity is pervasive in the following ways.  See
\cite{hall,Murtagh04}.

\begin{enumerate}
\item As dimensionality increases, so does ultrametricity.

\item 
In very high dimensional spaces, the ultrametricity approaches being 100\%.

\item 
Relative density is important: high dimensional and spatial sparsity mean 
the same in this context. 
\end{enumerate}

Compared to metric embedding, and model fitting in general, 
mapping metric (or other) data into an ultrametric, or embedding a 
metric in an ultrametric, leads to study of distortion.

As such a distortion, let us look at recoding of data, by modifying the data 
 precision.  By a focus on 
the data measurement process, we can find a new way to discover 
(hierarchical) structure in data.

\begin{figure}
\begin{center}
\includegraphics[scale=0.5]{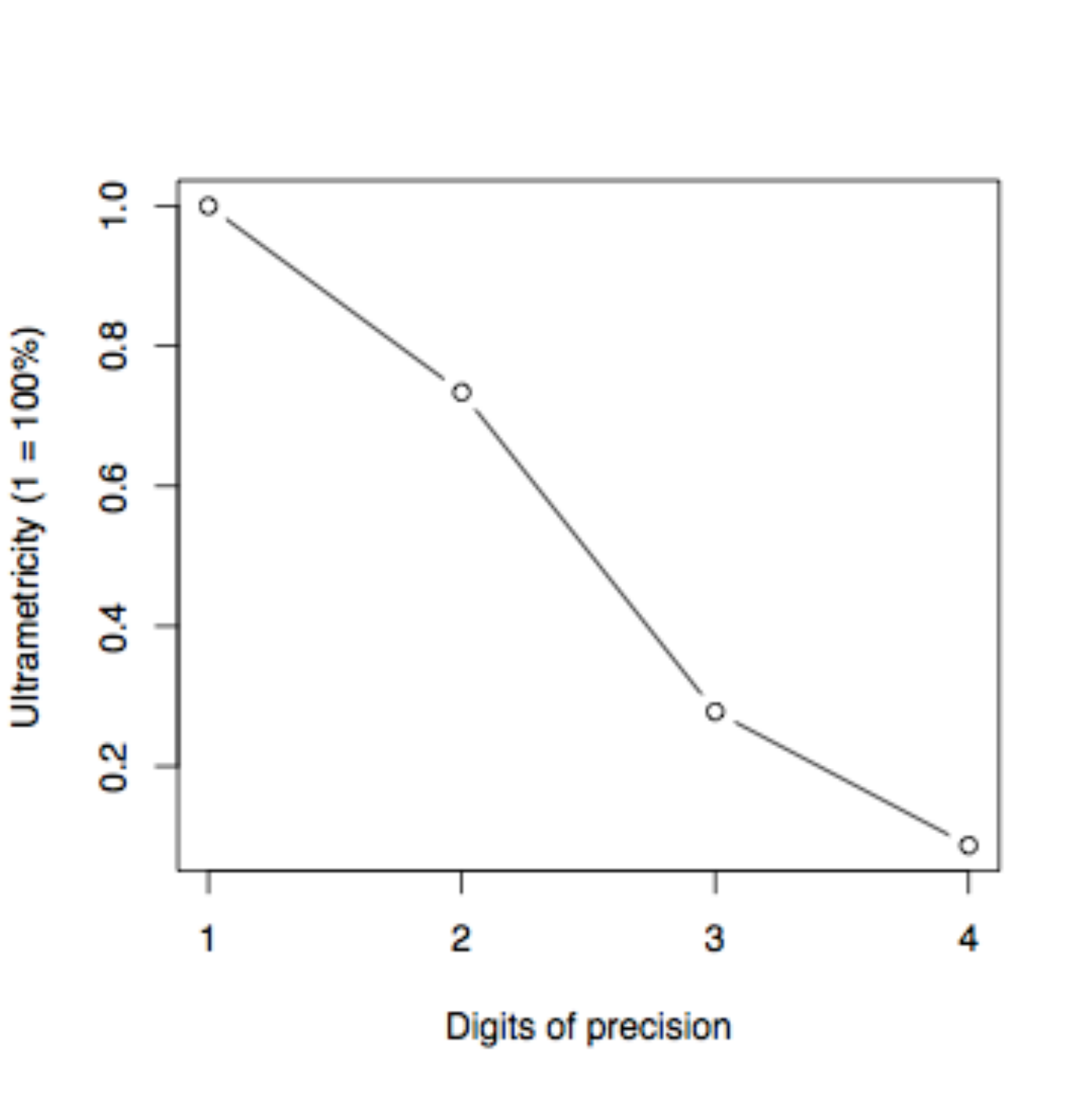}
\caption{Dependence of ultrametricity, i.e.\ data 
inherently hierarchical, on precision}
\label{figprecision}
\end{center}
\end{figure}

In Figure \ref{figprecision}, 20,000 encoded chemicals were used, 
normalized as described in \cite{Murtagh07}.   
Next, 2000 sampled triangles were selected, and ultrametricities obtained
for precisions $1, 2, 3, 4, \dots$  in all values.
Numbers of non-degenerate triangles (out of 2000) were found as follows
(where non-degenerate means isosceles with small base): 

\begin{itemize}
\item precision 1: 2
\item precision 2: 1062
\item precision 3: 1999
\item precision 4: 2000
\end{itemize}

Thus if we restrict ourselves to just 1 digit of precision we find 
a very high degree of ultrametricity, based -- to be noted -- on 
the preponderance of equilateral triangles.   With 2 digits of precision,
there are a lot more cases of isosceles triangles with small base.  

Thus we can bring about higher ultrametricity in a data set through 
reducing the precision of data measurement.  

\section{Generalized Ultrametric and Formal Concept Analysis}

The usual
ultrametric is an ultrametric distance, i.e.\ for a set I,
$d: I \times I \longrightarrow \R^+$. The range is the set of 
non-negative reals.  

The generalized
ultrametric is: $d: I \times I \longrightarrow \Gamma$,
where $\Gamma$ is a partially ordered set, or poset.  In other words, the
{\em generalized} ultrametric is a set.
The range of this generalized ultrametric is therefore defined on the power set
or join semilattice.  
The minimal element of the poset generalizes the 0 distance of 
the mapping onto $\R^+$ case.

Comprehensive background on ordered sets and
lattices can be found in \cite{davey}.
A review of generalized distances and ultrametrics
can be found in \cite{sedacj}.
Generalized ultrametrics are used in logic programming and, as we will
discuss in the subsection to follow, formal concept analysis can be seen as 
use of the generalized ultrametric.  

To illustrate how the generalized ultrametric is a vantage point on the Baire 
framework, we focus on the common, or shared, prefix aspect of this.  
Consider the following 
data values: $x_1 = 0.4573, x_2 = 0.4843, x_3 = 0.45635, x_4 = 0.4844, 
x_5 = 0.4504$.
Common prefixes are as follows, where we take decimal digits (i.e.\ ``4573''
in the case of $x_1$). 

\medskip

\noindent 
$d(x_1,x_2) = 4$

\noindent 
$d(x_1,x_3) = 4, 45$

\noindent 
$d(x_1,x_4) = 4$

\noindent 
$x(x_1,x_5) = 4, 45$

\noindent 
$d(x_2,x_3) = 4$

\noindent 
$d(x_2,x_4) = 4, 48, 484$

\noindent 
$d(x_2,x_5) = 4$

\noindent 
$d(x_3,x_4) = 4$

\noindent 
$d(x_3,x_5) = 4, 45$

\noindent 
$d(x_4,x_5) = 4$

\medskip

The partially ordered set is just the structure with the ``subset of the common prefix'' 
binary relation with, on one level, 
the single valued 
common prefixes (here: 4); the next level has the common prefixes of length 2 (here:
45, 48); the following level has the common prefixes of length 3 (here: 484).  Prior to the
first level, we have the length 0 common prefix corresponding to no match between 
the strings.  

Thus we see how we can read off a partially ordered set, contained in a lattice, 
based on the common or shared prefixes.  

\subsection{Formal Concept Analysis}

In formal concept analysis (FCA) \cite{davey}, we focus on the lattice 
as described in the previous subsection.
Thus, we can say that a lattice representation is yet another way of 
displaying (and structuring) the clusters and the cluster assignments 
in our Baire framework.  

Following a formulation by Mel Janowitz (see \cite{jan0,jan1}), lattice-oriented
FCA is contrasted with poset-oriented hierarchical clustering in the following 
way, where by ``summarize'' is meant that the data structuring through 
lattice or poset allows statements 
to be made about the cluster members or other cluster properties.  

\begin{itemize}
\item 
Cluster, then summarize.  This is the approach taken by (traditional) hierarchical clustering.
\item 
Summarize, then cluster. This is, in brief, the approach taken by FCA.
\end{itemize}

Further description of FCA is provided in \cite{CMJoC}.  Our aim here has been 
to show how the common prefixes of strings leads to the Baire distance, 
and also to a generalized ultrametric, and furthermore to a poset and 
an embedding in a lattice.  

\section{Linear Time and Direct Reading Hierarchical Clustering}

\subsection{Linear Time, or $O(N)$ Computational Complexity, Hierarchical
Clustering}
\label{subsect81}

A point of departure for our work has been the computational objective
of bypassing computationally demanding hierarchical clustering 
methods (typically quadratic time, or $O(N^2)$ for $N$ input 
observation vectors), but also having a framework that is 
of great practical importance in terms of the application domains.

Agglomerative hierarchical clustering algorithms are based on 
pairwise distances (or dissimilarities) implying computational 
time that is $O(N^2)$ where $N$ is the number of observations.
The implementation required to achieve this is, for most agglomerative
criteria, the nearest neighbor chain, together with the reciprocal
nearest neighbors, algorithm
(furnishing inversion-free 
hierarchies whenever Bruynooghe's reducibility property, see \cite{Murtagh85-1}, 
is satisfied by the cluster criterion). 

This quadratic time requirement is a worst case performance result.   
It is most often the average time also since the pairwise agglomerative 
algorithm is applied directly to the data without any preprocessing 
speed-ups (such as preprocessing that facilitates fast nearest neighbor 
finding).   An example of a linear average time algorithm for 
(worst case quadratic computational time) agglomerative hierarchical
clustering is in \cite{Murtagh83ipl}.

With the Baire-based hierarchical clustering algorithm, we have an algorithm
 for linear time worst case hierarchical clustering.  It can be characterized
as a divisive rather than an agglomerative algorithm.
  
\subsection{Grid-Based Clustering Algorithms}
\label{subsection:grid-based-clustering}

The Baire-based hierarchical clustering algorithm has characteristics
that are related to grid-based clustering algorithms, and density-based clustering
algorithms, which -- often -- were developed in order to handle very large 
data sets.   

The main idea here is to use a grid like structure to split the information space, separating the dense grid regions from the less dense ones to form groups.
In general, a typical approach within this category will consist of the following steps~\cite{Grabusts02}:
\begin{enumerate}
        \item Creating a grid structure, i.e.\ partitioning 
    the data space into a finite number of  non-overlapping cells.
        \item Calculating the cell density for each cell.
        \item Sorting of the cells according to their densities.
        \item Identifying cluster centers.
        \item Traversal of neighbor cells.
\end{enumerate}

Additional information about grid-based clustering can be 
found in the following works~\cite{Chang02, Gan07, Park04, Xu08}.

In sections \ref{sect4} and \ref{hashing} in particular it has been 
described 
how cluster bins, derived from an m-adic tree, provide us with 
a grid-based framework or data structuring.  We can read off the 
cluster bin members from such an m-adic tree.  In subsection 
\ref{subsect81} we have noted how an m-adic tree requires one 
scan through the data, and therefore this data structure is 
constructed in linear computational time.  

In such a preprocessing context, clustering with the Baire distance 
can be seen as a ``crude'' method 
for getting clusters.
After this we can use more traditional techniques
to refine the clusters in terms of their membership.  
Alternatively (and we have quite extensively compared Baire clustering 
with, e.g.\
k-means, where it compares very well) clustering with the Baire distance 
can be seen as fully on a par with any optimization algorithm for 
clustering.  As optimization, and just as one example from the many 
examples reviewed in this article, the Baire approach optimizes
an m-adic fit ot the data simple by reading the m-adic structure
directly from the data.

\section{Conclusions: Many Viewpoints, Various Implementations}

Baire distance is an ultrametric, so we can think of reading off 
observations as a tree.
 
Through data precision of measurement, alone, we can enhance inherent 
ultrametricity, or inherent hierarchical properties in the data.  

 Clusters in such a Baire-based hierarchy are simple ``bins'' and 
assignments are determined through a very simple hashing.   (E.g.\ 
$0.3475 \longrightarrow$ bin 3, and $\longrightarrow$ bin 34, and 
$\longrightarrow$ bin 347, and $\longrightarrow$ bin 3475.)

 Observations are mapped onto sets.   (E.g.\ 0.3475 and 0.3462 are 
mapped onto sets labeled by 3 and 34.)  We have therefore a 
generalized ultrametric.   A lattice can be used to represent 
range sets, leading also to a link with formal concept analysis.   

Our wide range of vantage points on the Baire-based processing is 
important because of the many, diverse applications in view,  
including the structuring of the data, the reading off of clusters, the 
matching or best fit of observations, the determining of hierarchical properties,
and so on.   

Apart from showing how the Baire vantage point gives rise in practice to 
such a breakthrough result of having linear time hierarchical clustering our 
other important contribution in this work has been to show how so many 
vantage points can be adopted in this way on data, on the structuring and 
embedding of data, and ultimately on the interpretation and exploitation of
data.  

\bibliographystyle{plain}
\bibliography{bibliography}

\end{document}